\documentclass{article}

    \usepackage[final]{tackling_climate_workshop_style}

\usepackage[utf8]{inputenc} 
\usepackage[T1]{fontenc}    
\usepackage{hyperref}       
\usepackage{url}            
\usepackage{booktabs}       
\usepackage{amsfonts}       
\usepackage{nicefrac}       
\usepackage{microtype}      
\usepackage{amsmath}
\usepackage{graphicx}
\usepackage{wrapfig}        
\usepackage[dvipsnames]{xcolor}
\usepackage{subcaption}
\usepackage{booktabs}
\usepackage{siunitx}
\usepackage{algpseudocode}
\usepackage{algorithm}

\usepackage{todonotes}

\title{Sensitivity Analysis for Climate Science with Generative Flow Models}

\begin{document}
\author{
  Alex Dobra\thanks{Corresponding author: alex.dobra@robots.ox.ac.uk}\\
  Dept. of Engineering\\
  University of Oxford\\
  \And
  Jakiw Pidstrigach \\
  Dept. of Statistics \\
  University of Oxford\\
  \And
Tim Reichelt \\
Dept. of Physics\\
University of Oxford \\
  \AND
Paolo Fraccaro \\
IBM\\
  \And
Johannes Jakubik \\
IBM\\
  \And
Anne Jones \\
IBM\\
  \And
Christian Schroeder de Witt \\
Dept. of Engineering\\
University of Oxford \\
  \And
Philip Torr \\
Dept. of Engineering\\
University of Oxford \\
  \And
Philip Stier \\
Dept. of Physics\\
University of Oxford \\
}

\maketitle

\begin{abstract}
    Sensitivity analysis is a cornerstone of climate science, essential for understanding phenomena ranging from storm intensity to long-term climate feedbacks. However, computing these sensitivities using traditional physical models is often prohibitively expensive in terms of both computation and development time. While modern AI-based generative models are orders of magnitude faster to evaluate, computing sensitivities with them remains a significant bottleneck. This work addresses this challenge by applying the adjoint state method for calculating gradients in generative flow models. We apply this method to the cBottle generative model, trained on ERA5 and ICON data, to perform sensitivity analysis of any atmospheric variable with respect to sea surface temperatures. We quantitatively validate the computed sensitivities against the model's own outputs. Our results provide initial evidence that this approach can produce reliable gradients, reducing the computational cost of sensitivity analysis from weeks on a supercomputer with a physical model to hours on a GPU, thereby simplifying a critical workflow in climate science. The code can be found at \url{https://github.com/Kwartzl8/cbottle_adjoint_sensitivity}.
\end{abstract}

\section{Introduction}
Sensitivity analysis is widely applied throughout climate science, e.g. for investigating the drivers of storms \cite{banomedina2025Xynthia}, the importance of observations \cite{torn2008ensemble} or understanding climate feedbacks \cite{bloch2024green,van2025reanalysis}. However, conducting sensitivity analysis with classical physical simulation models is prohibitively expensive, relying either on deriving and implementing adjoints by hand \cite{errico1997adjoint,tremolet2007model,doyle2014initial} or running finite difference approximations \cite{bloch2024green}. 
For example, a standard approach for estimating the response of the atmosphere to spatial changes in temperature fields relies on Green's function method which requires executing an expensive General Circulation Model (GCM) for thousands of model years which can take weeks, even on a supercomputer \cite{branstator1985analysis,holzer2000transit,barsugli2002global,zhou2017analyzing,dong2019attributing,bloch2024green,van2025reanalysis}.

Recently, there has been an explosion of development of large AI models for weather and climate. While expensive to train, they are significantly faster and differentiable by default, warranting an investigation into their potential to simplify the sensitivity analysis workflow. Modern AI models are developed in frameworks like PyTorch \cite{pytorch} and JAX \cite{jax2018github}, which are equipped with automatic differentiation (AD) engines \cite{baydin2018automatic}. Consequently, obtaining the gradient of any scalar quantity derived from the model's output with respect to any model input only requires a single function call to the AD engine, which has motivated their use for initial condition sensitivity studies \cite{banomedina2025Xynthia,vonich2024predictability,toride2025using}.

\begin{figure}[h]
    \centering
    \begin{subfigure}[c]{0.44\textwidth}
        \includegraphics[width=\linewidth]{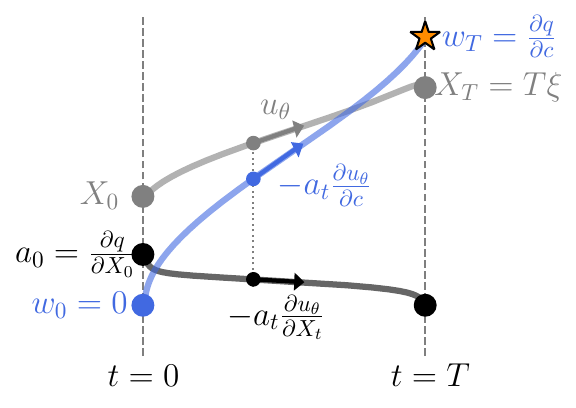}
    \end{subfigure}
    \hfill
    \begin{subfigure}[c]{0.55\textwidth}
        \includegraphics[width=\linewidth]{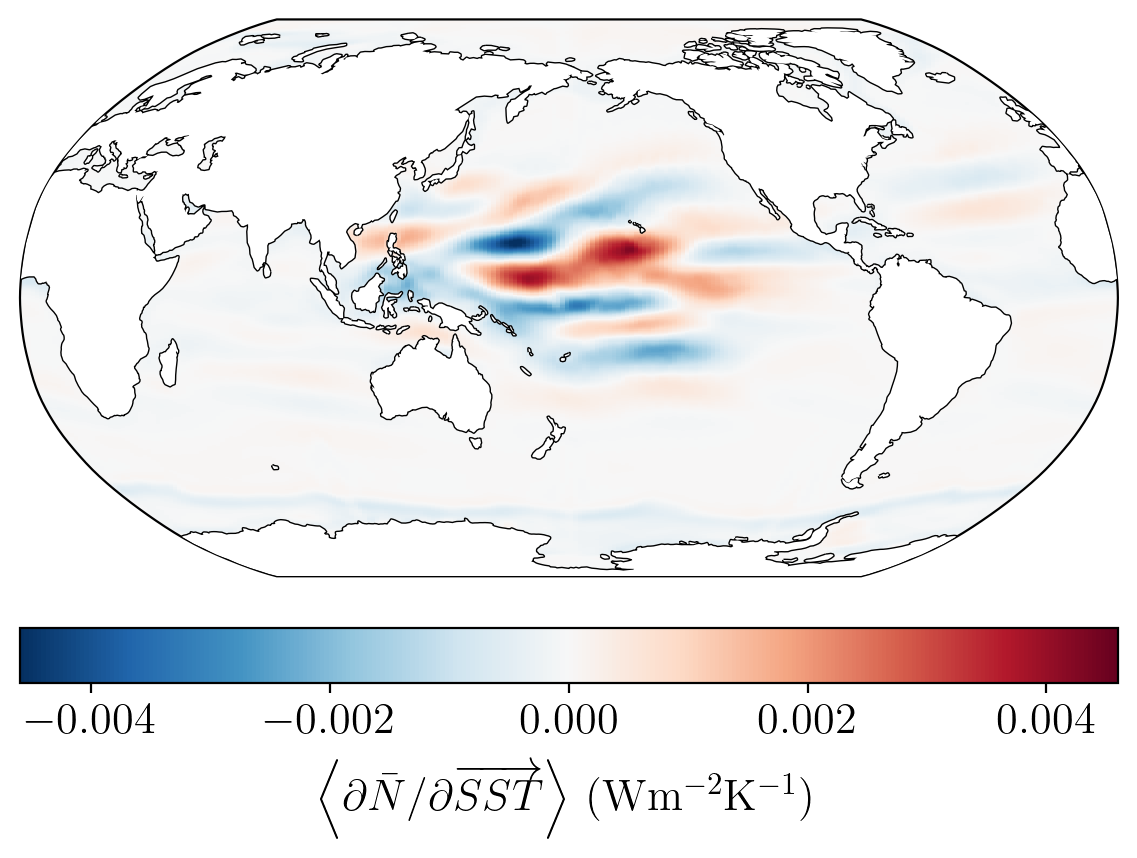}
    \end{subfigure}
    \caption{\textbf{Left.} An illustration of obtaining the sensitivity $\frac{\partial q}{\partial c}$ with the adjoint method. The velocity network $u_{\theta}(X_t,t,c)$ turns the clean vector $X_0$ into a scaled noise latent $X_T$. As $u_{\theta}$ is steered by conditioning $c$ throughout the reverse sampling process, we need a variable $w_t$ which accumulates the gradient from all noise levels. The gradient accumulator depends on the adjoint $a_t=\frac{\partial q}{\partial X_0} \frac{\partial X_0}{\partial X_t}$, which expresses the sensitivity of $q$ with respect to a partially noised vector $X_t$. Integrating their respective ODEs together yields $\frac{\partial q}{\partial c}=w_T$. \textbf{Right.} Net global radiation flux sensitivity to SST obtained from cBottle, averaged over historical AMIP SST 1971-2020.
    }
    \label{fig:adjoint_illustration}
\end{figure}
Notably, flow and diffusion models \cite{sohl2015deep,ho2020denoising,song2021scorebased,lipman2023flow} have gained popularity in weather and climate due to their powerful generative properties and their ability to model uncertainty. Diffusion models have demonstrated state-of-the-art results in weather forecasting \cite{price2025probabilistic} and are able to reproduce atmospheric states \cite{brenowitz2025climatebottlegenerativefoundation}. The method presented in this paper provides a way to extract gradients from these types of models, reducing the cost of sensitivity analysis from days to weeks on a supercomputer to minutes or hours on a GPU. However, it is not as clear if these models can produce informative gradients even if they provide a good fit to the data. To begin the work of evaluating their quality, this paper introduces a method of extracting gradients from flow models and checking their consistency with model predictions through finite differences.


\section{Background and notation}

\textbf{Climate in a Bottle.} The experiments in this work are done with the simplest Climate in a Bottle diffusion model \cite{brenowitz2025climatebottlegenerativefoundation} (henceforth called cBottle). Given the day of year $\tau$, second of day $\zeta$, sea surface temperature forcing $c$, and a high dimensional random vector $\xi$, the cBottle model generates 45 atmospheric variables ($4$ 3D-variables at $8$ pressure levels and $13$ 2D-variables) at a ground resolution of $\sim100~\text{km}$. See App. \ref{app:cBottle} for details. cBottle has been trained on both ERA5 reanalysis data \cite{hersbach2020era5} and outputs from the ICON climate model \cite{hohenegger2023icon}, providing a unique test-bed for gradient extraction methods.
While the empirical results are computed with cBottle, the method presented here is not specific to cBottle and applies to generative flow models in general.

\textbf{Flow models.} Diffusion models \cite{sohl2015deep, ho2020denoising, song2021scorebased}, and the more recent flow matching \cite{lipman2023flow, liu2022flowstraightfastlearning, albergo2023stochastic} are all special cases of flow models, a family of generative models which learn a time-varying velocity field $u_\theta(X_t, t)$ that transports an easy-to-sample probability distribution $p_{init}$ to a data probability distribution $p_{data}$. In what follows, we will use the parameterization of the Elucidated Diffusion Model (EDM) \cite{karras2022edm} with deterministic sampling.

\textbf{Conditioning.} Flow models can be guided by input variables $c$ to sample from the conditional distribution $p_{data}\left( \cdot | c \right)$ \cite{zhang2024control_net,denker2024deft,pidstrigach2025conditioning}. Therefore, the procedure for generation implies sampling $\xi$ from the standard Gaussian, picking a conditioning variable $c$ and solving the ODE
$$\text dX_t=u_\theta(X_t, t, c)\text dt, \text{ with } X_{T}=T\xi \text{ and }\xi \sim \mathcal{N}\left(0,\text{Id}\right)$$
backwards in time, from $T$ to $0$, obtaining the generated sample $X_0\sim p_{data}\left( \cdot | c \right)$.

\section{Method}

\begin{figure}
    \centering
    \begin{subfigure}{0.43\textwidth}
        \includegraphics[width=\linewidth]{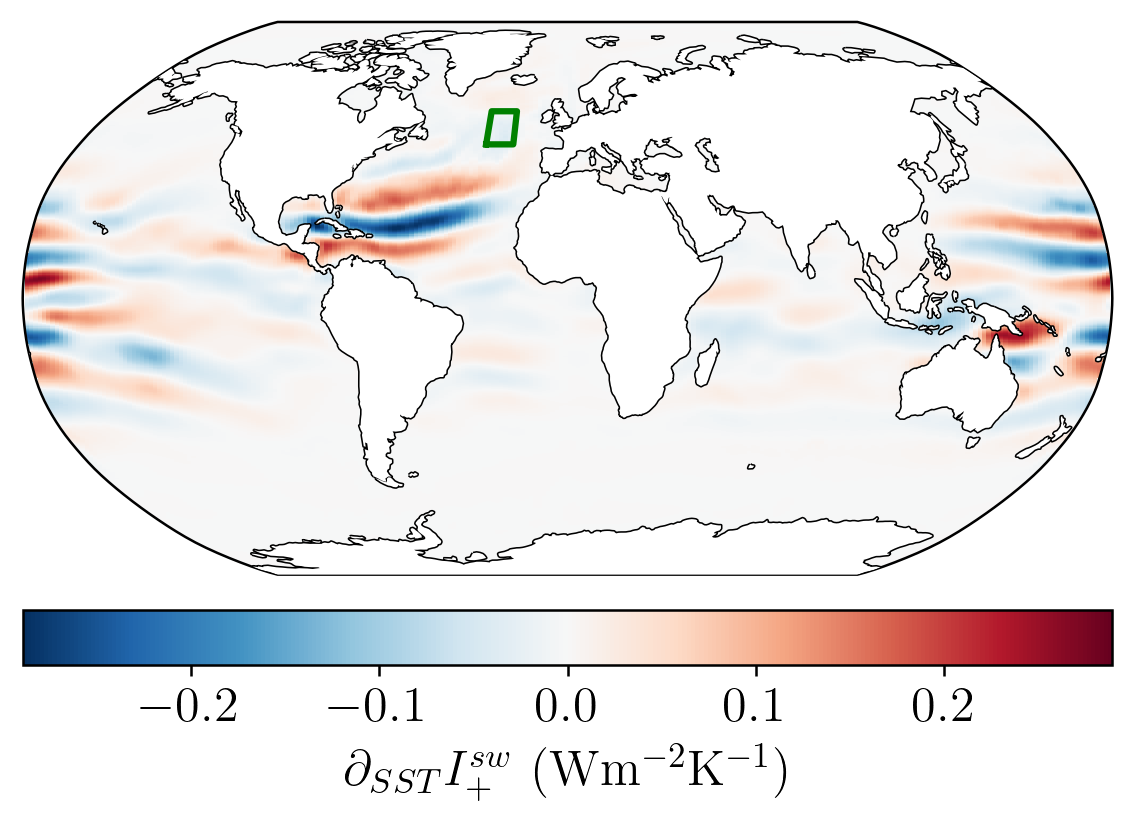}
    \end{subfigure}
    \hfill
    \begin{subfigure}{0.56\textwidth}
        \includegraphics[width=\linewidth]{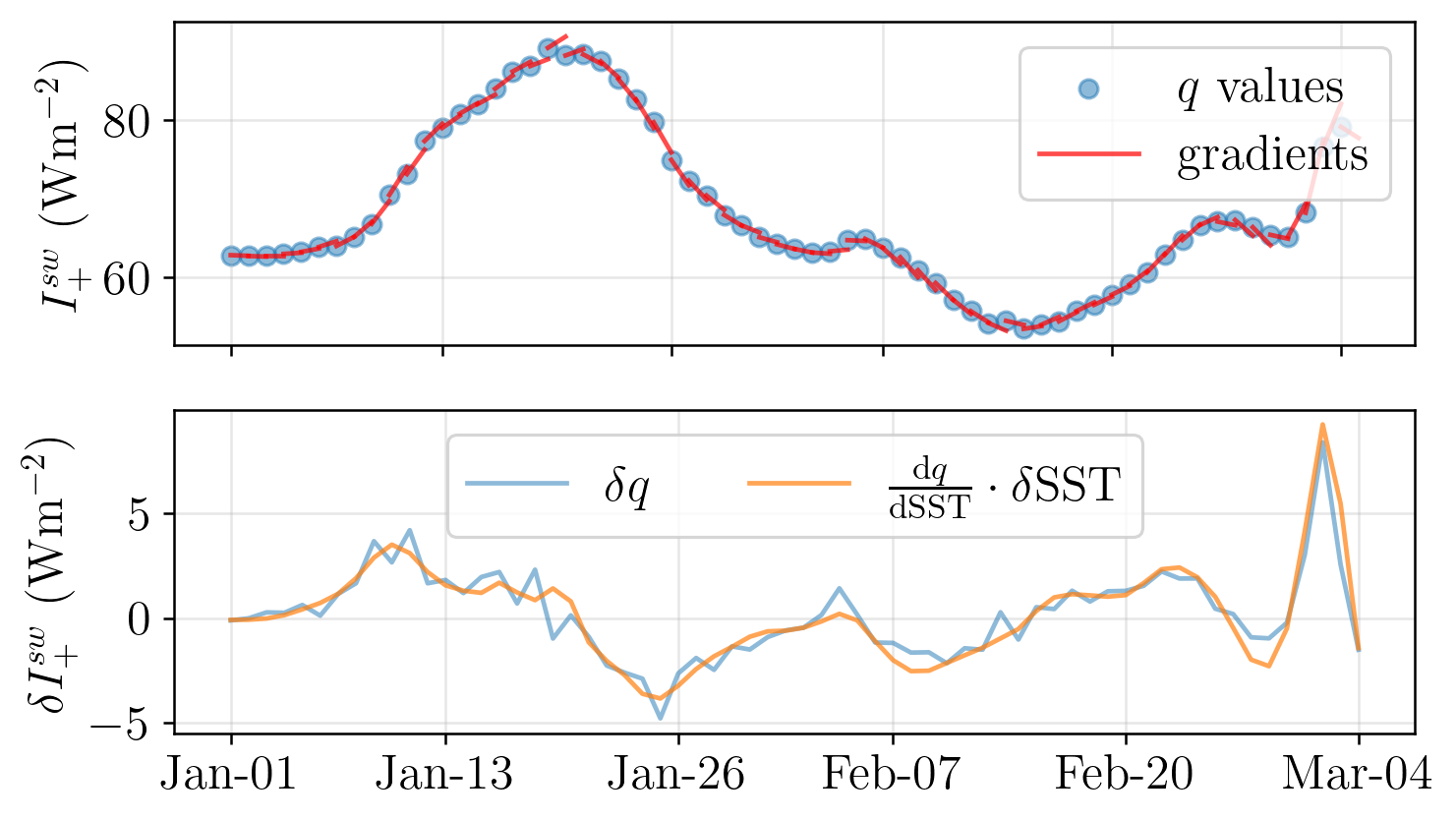}
    \end{subfigure}
    \caption{\textbf{Left.} A sensitivity map $\frac{\partial q}{\partial \text{SST}}$ on February 1st 2020, at 17:30 UTC. Here, $q$ is the average outgoing shortwave radiation in the green patch $I_+^{sw}$. \textbf{Right.} The top plot shows how $q$ changes over the course of two months, and the extracted gradients. The bottom plot shows the true finite difference $\delta q$ and the linearized difference $\frac{\text dq}{\text{dSST}}\cdot \delta \text{SST}$, see equation \ref{eq:cbottle_grad_check}. The RMSE (equation \ref{eq:rmse_metric}) between the two is $0.76$ Wm$^{-2}$.
    }
    \label{fig:q_grad_check}
\end{figure}

We are interested in a scalar quantity derived from a generated sample $q(X_0)$ and how this quantity varies with conditioning $\frac{\partial q}{\partial c}$. For example, in the context of cBottle, $q$ can be net outgoing radiation globally, precipitation on the coast of Peru or wind shear in the tropical Pacific, and $c$ is global sea surface temperatures. As flow models have very deep computational graphs due to the recurrent calls to the model, it is not feasible to simply use AD to obtain this derivative. We first write down a continuous counterpart of the chain-rule
$$
\frac{\partial q}{\partial c}=\frac{\partial q}{\partial X_0}\frac{\partial X_0}{\partial c}=\frac{\partial q}{\partial X_0}\int_{T}^0 \frac{\partial X_0}{\partial X_t} \frac{\partial u_{\theta}}{\partial c}(X_t, t, c)\text dt.
$$

The key step is to use the adjoint state $a_t=\frac{\partial q}{\partial X_0} \frac{\partial X_0}{\partial X_t}$, which satisfies the ODE
$$
\frac{\text{d}}{\text{d}t}a_t=-a_t\frac{\partial u_{\theta}}{\partial X_t}.
$$
Letting $w_{T-t}=\int^t_T a_s\frac{\partial u_{\theta}}{\partial c}\text ds$ we see the desired result is $\frac{\partial q}{\partial c}=a_0\frac{\partial X_0}{\partial c}=w_T$. Solving the system of ODEs
\begin{equation}\label{eq:adjoint_method}
\frac{\text{d}}{\text{d}t}
\begin{bmatrix}
X_t\\
a_t\\
w_t
\end{bmatrix} = 
\begin{bmatrix}
u_{\theta}\\
-a_t\frac{\partial u_{\theta}}{\partial X_t}\\
-a_t\frac{\partial u_{\theta}}{\partial c}
\end{bmatrix}, \text{ with }
\begin{bmatrix}
X_0\\
a_0\\
w_0
\end{bmatrix} = 
\begin{bmatrix}
X_0\\
\frac{\partial q}{\partial X_0}\\
0
\end{bmatrix},
\end{equation}
from $0$ to $T$ yields $\frac{\partial q}{\partial c}=w_T$. The process is depicted in Figure~\ref{fig:adjoint_illustration}, and proofs can be found in \cite{pontryagin1962optimal,chen2018neuralODEs,schurov2022adjoint}.

\section{Results with Climate in a Bottle}
We conduct our experiments with cBottle, a diffusion model of the EDM type \cite{karras2022edm}. It can be easily reparameterized as a flow model $\text dX_t=u_\theta(X_t, t, c, \tau, \zeta)\text dt, \text{ with } X_{T}=T\xi \text{ and }\xi \sim \mathcal{N}\left(0,\text{Id}\right)$ via the probability flow ODE \cite{song2021scorebased}.
The conditioning variables are $c$ -- sea surface temperature (SST) forcing, $\tau$ -- day of the year and $\zeta$ -- UTC second of day. We generate samples from scratch given the conditioning and a randomly drawn noise sample, and then we apply the adjoint sensitivity algorithm with the same conditioning. For the gradient self-consistency checks, we keep the noise samples $\xi$ constant.

\textbf{Gradient self-consistency check.} The approximate nature of the flow model mapping from $c$ to $X_0$ means the reliability of its gradients is not immediately clear. While the model's outputs are typically validated for consistency against real-world data or physical models \cite{brenowitz2025climatebottlegenerativefoundation}, the trustworthiness of the gradients is commonly not assessed. We bridge this gap by proposing a self-consistency check that validates the model's gradients against its own outputs, thereby using the validation of the predictive samples to establish confidence in the model gradients. Figure~\ref{fig:q_grad_check} shows the gradient self-consistency check procedure for a selected quantity $q$ chosen to be the average outgoing shortwave radiation $I_+^{sw}$ in a small area in the north Atlantic. The gradients obtained through the adjoint sensitivity method seem consistent with cBottle's outputs. To quantify this, we define a simple RMSE metric over $K$ samples, following \cite{bloch2024green}:

\begin{equation}\label{eq:rmse_metric}
    \text{RMSE}=\sqrt{\frac{1}{K}\sum_k\left( \delta q - \frac{\text dq}{\text{d}c}\cdot \delta c \right)^2},
\end{equation}
where the total linearized difference is obtained by the chain rule:
\begin{equation}\label{eq:cbottle_grad_check}
\frac{\text dq}{\text{d}c}\cdot \delta c = \frac{\partial q}{\partial c}\cdot \delta c+\frac{\partial q}{\partial \tau}\delta\tau+\frac{\partial q}{\partial \zeta}\delta\zeta.    
\end{equation}
The derivatives with respect to the other conditioning variables are simply obtained by setting up gradient accumulators for each one, see Appendix \ref{app:full_derivative} for details.

\textbf{The sensitivity of global net radiative flux with respect to SST patterns.} While the gradient self-consistency check confirms the correctness of the algorithm, it does not guarantee physical meaning. To test this, we follow \cite{bloch2024green} (henceforth called GFMIP) and calculate the sensitivity of average global net radiative flux with respect to SST $\frac{\partial \bar N}{\partial \text{SST}}$, averaged over $K=2536$ samples of SST in the AMIP dataset from 1971-2020. For this run, the noise latent $\xi$ was sampled from the standard Gaussian. The timestep was set to \qty{169}{h}, which is 1 week and 1 hour, to uniformly sample the two time conditioning variables in cBottle, $\tau$ and $\zeta$. The resulting sensitivity map is presented in the right panel of Figure~\ref{fig:adjoint_illustration}. The RMSE for the entire period is $\qty{0.465}{Wm^{-2}}$, and if we take the RMSE of the yearly averaged differences like in GFMIP, we get $\qty{0.07}{Wm^{-2}}$, which is close to their value, $\qty{0.23}{Wm^{-2}}$. GFMIP do not have high spatial frequencies like ours, but this is to be expected because the patch perturbations used are much larger than the $\sim \qty{100}{km}$ pixels in cBottle and they smooth out the pattern. There are clear negative feedbacks around the Maritime Continent, which might be explained by an increase in convective clouds and thus reduced outgoing long-wave radiation. A faint ring-shaped negative signal around the Antarctic ice-sheet can also be observed, which can be explained by a decrease in global albedo caused by ice-sheet melting. However, the sensitivity in the Pacific is 1-2 orders of magnitude larger than in GFMIP, even as the signals in the other oceans are around the same order of magnitude.

\section{Discussion and future work}
\textbf{Limitations.} In the case of the sensitivity of global net radiation with respect to SST patterns, the large departure from previous results warrants further investigation. It is possible that cBottle's outputs are very strongly conditioned on the day of year $\tau$, which implies that simply averaging $\frac{\partial q}{\partial c}$ is not the full picture and the procedure described in Appendix \ref{app:full_derivative} must be followed to add the contribution of the $\frac{\partial q}{\partial \tau}$ gradients.

It must also be said that there is no guarantee cBottle produces physical atmospheres for out-of-distribution SST values (for example $+\qty{2}{K}$). In fact, we noticed tiling artifacts appearing for gradients with respect to climatology SST when trying to recreate GFMIP's results.

\textbf{Future work.} A promising avenue for greatly extending the applications of this method is model guidance. To condition the model to generate desired weather states from the exact posterior (for example, show a realistic hurricane in the Atlantic), we can train a separate model \cite{zhang2024control_net,denker2024deft,pidstrigach2025conditioning} that uses a guiding variable $y=G(X_0)$, where $G$ is an observation operator. Gradients with respect to $y$ can then be pulled through the small guiding model. This method would allow calculations of gradients of anything with respect to anything, including variables not included in the original model, for example CO$_2$ concentrations, aerosol optical depths or ocean salinity.

\section*{Acknowledgements}
AD is part of the Intelligent Earth CDT supported by funding from the UK Research and Innovation Council (UKRI) grant number EP/Y030907/1. JP acknowledges support from EPSRC (grant number EP/Y018273/1). TR is supported by the EU's Horizon Europe program under grant agreement number 101131841 and also received funding from UK Research and Innovation (UKRI).

AD thanks Milan Kloewer for the insightful discussions on the physical meaning of extracted sensitivities, Lilli Freischem for an introduction to regridding and Peter Manshausen for feedback on early results.

\bibliographystyle{unsrt}
\bibliography{references}

\appendix
\section{Getting total derivatives with the adjoint sensitivity method}\label{app:full_derivative}
In the case in which the inputs to the flow model are dependent on each other, as is the case with cBottle's forcing SSTs $c$ and day of year $\tau$, we might be interested in total derivatives, in addition to the partial derivatives obtainable through the adjoint sensitivity method. To reiterate, obtaining partial derivatives (from a velocity-reparameterized EDM) with the adjoint sensitivity method implies solving the system of ODEs
$$
\frac{\text{d}}{\text{d}t}
\begin{bmatrix}
X_t\\
a_t\\
w_t\\
v_t
\end{bmatrix} = 
\begin{bmatrix}
u_{\theta}\\
-a_t\frac{\partial u_{\theta}}{\partial X_t}\\
-a_t\frac{\partial u_{\theta}}{\partial c}\\
-a_t\frac{\partial u_{\theta}}{\partial \tau}
\end{bmatrix}, \text{ with }
\begin{bmatrix}
X_0\\
a_0\\
w_0\\
v_0
\end{bmatrix} = 
\begin{bmatrix}
X_0\\
\frac{\partial q}{\partial X_0}\\
0\\
0
\end{bmatrix}
$$
from $t=0$ to $t=T$, where $w_T=\frac{\partial q}{\partial c}$ and $v_T=\frac{\partial q}{\partial \tau}$. In practice, calculating the additional derivative $\frac{\partial u_{\theta}}{\partial \tau}$ adds negligible computational cost, as the backpropagation process through the computational graph happens the same way and with the same starting tangent $-a_t$. Any input of the model can be added to the ODE this way. The total derivative is then
$$
\frac{\text dq}{\text dc}=\frac{\partial q}{\partial c}+\frac{\partial q}{\partial \tau}\frac{\partial \tau}{\partial c}=w_T+v_T\frac{\partial \tau}{\partial c}.
$$
To approximate $\frac{\partial \tau}{\partial c}$, one could train a regressor and get this gradient with an AD call.

For calculating the RMSE on historical data this is not needed, as we have both $\delta c$ and $\delta \tau$:
$$
\frac{\text dq}{\text dc}\cdot \delta c=\frac{\partial q}{\partial c}\cdot \delta c+\frac{\partial q}{\partial \tau}\frac{\partial \tau}{\partial c}\cdot \delta c\approx w_T\cdot \delta c+v_T\delta \tau.
$$

\section{More sensitivity maps}
\begin{figure}
    \centering
    \includegraphics[width=\linewidth]{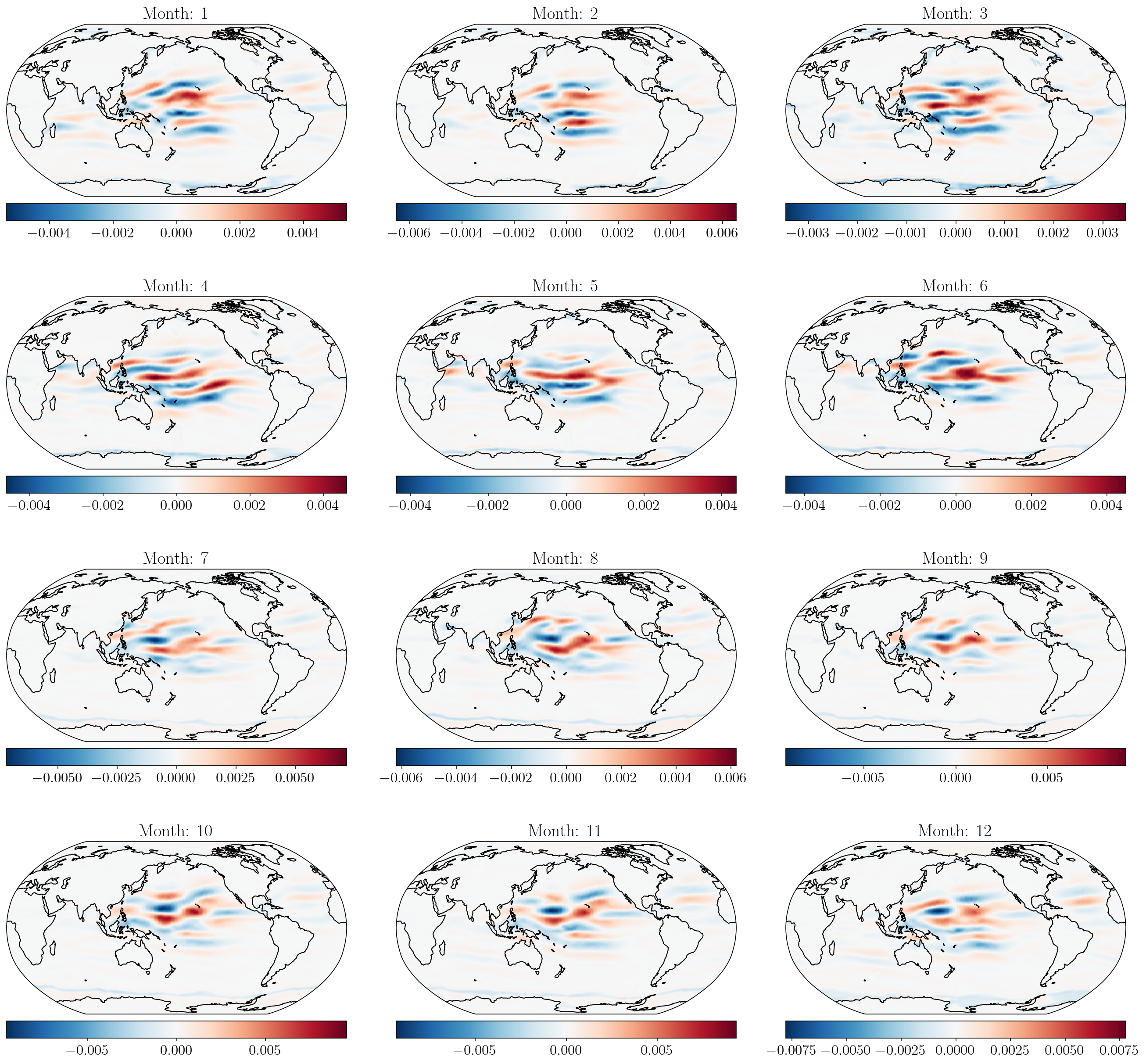}
    \caption{Monthly averaged sensitivity maps $\frac{\partial \bar N}{\partial \text{SST}}$ in the period 1971-2020.}
    \label{fig:monthly_sensitivity_maps}
\end{figure}

See Figure~\ref{fig:monthly_sensitivity_maps} for monthly sensitivity maps, each one averaged over $K/12\approx211$ samples.

\section{cBottle details}\label{app:cBottle}

\subsection{Generated variables}\label{app:cBottle_variables}
See Table \ref{table:cbottle_vars}.
\begin{table}[ht]
\centering
\begin{tabular}{lll}
\toprule
\textbf{Short name} & \textbf{Description} & \textbf{Units} \\
\midrule
\multicolumn{3}{c}{\textbf{Profile variables} @ 1000, 850, 700, 500, 300, 200, 50, 10 \si{hPa}} \\
\midrule
\verb|T| & air temperature (profile) & \si{K} \\
\verb|U| & zonal wind (profile) & \si{m.s^{-1}} \\
\verb|V| & meridional wind (profile) & \si{m.s^{-1}} \\
\verb|Z| & geopotential height & \si{m} \\
\midrule
\multicolumn{3}{c}{\textbf{2D variables}} \\
\midrule
\verb|clivi| & column integrated cloud ice & \si{kg.m^{-2}} \\
\verb|cllvi| & column integrated cloud water & \si{kg.m^{-2}} \\
\verb|rlut|  & TOA outgoing longwave radiation & \si{W.m^{-2}} \\
\verb|rsut|  & TOA outgoing shortwave radiation & \si{W.m^{-2}} \\
\verb|rsds|  & surface downwelling shortwave radiation & \si{W.m^{-2}} \\
\verb|tcwv|  & column integrated water vapour & \si{kg.m^{-2}} \\
\verb|pr|   & precipitation flux & \si{kg.m^{-2}.s^{-1}} \\
\verb|pres_msl| & mean sea level pressure & \si{Pa} \\
\verb|sic|  & sea ice concentration (fractional) & -- \\
\verb|sst|  & sea surface temperature & \si{K} \\
\verb|tas|  & 2m air temperature & \si{K} \\
\verb|uas|  & zonal wind at 10m & \si{m.s^{-1}} \\
\verb|vas|  & meridional wind at 10m & \si{m.s^{-1}} \\
\bottomrule
\vspace{0pt}
\end{tabular}
\caption{Variables grouped into profile, column, and surface categories with their ICON names, descriptions, and units.}
\label{table:cbottle_vars}
\end{table}
\subsection{The nature of SST conditioning data}\label{app:sst_nature}
Climate in a Bottle is using one of its conditioning variables, the day of year $\tau$, to obtain the sea surface temperature forcing $c(\tau)$. This procedure is mentioned but not detailed in their paper, and so we will describe here. They use the AMIP dataset \verb|input4MIPs.CMIP6Plus.CMIP.PCMDI.PCMDI-AMIP-1-1-9.ocean.mon.tosbcs.g| which can be downloaded with a script \href{https://esgf.ceda.ac.uk/thredds/dodsC/esg_cmip6/input4MIPs/CMIP6Plus/CMIP/PCMDI/PCMDI-AMIP-1-1-9/ocean/mon/tosbcs/gn/v20230512/tosbcs_input4MIPs_SSTsAndSeaIce_CMIP_PCMDI-AMIP-1-1-9_gn_187001-202212.nc}{from here}. This dataset contains the estimated global mid-month, monthly mean SST from January 1870 to December 2022, all dated on the 16th of the month (except February, which has the 15th). If the model is given a day of year $\tau$, $c(\tau)$ is linearly interpolated from the nearest two data points:
$$
c(\tau) = c(\tau_{i})+ \frac{\tau - \tau_i}{\tau_{i+1}-\tau_i}\left(c(\tau_{i+1})-c(\tau_{i})\right),
$$
where $\tau_i$ is the mid-month day-of-year of month $i$ and $\tau$ satisfies $\tau_i\leq\tau<\tau_{i+1}<\tau+31$.

For simplicity, we have maintained that $\tau$ is the day of the year, but it would be more precise to say it is the day of the year plus the day fraction: $\tau=1.5$ means January 1st 12:00 UTC.


\end{document}